\documentclass{article}

\usepackage[final]{nips_2018}

\usepackage[english]{babel}
\usepackage[utf8x]{inputenc}
\usepackage[T1]{fontenc}
\usepackage[noend]{algpseudocode}
\usepackage{fancyhdr,amssymb}
\usepackage[ruled,vlined]{algorithm2e}
\usepackage{array}
\usepackage{stmaryrd}
\usepackage{amsfonts}

\usepackage{amsfonts,amsmath,amssymb,amsthm,epsfig,psfrag}
\usepackage[T1]{fontenc}
\usepackage{babel}

\usepackage{amsmath}
\usepackage{mathtools}
\usepackage{dsfont}
\usepackage{systeme}
\usepackage{graphicx}
\usepackage[colorinlistoftodos]{todonotes}
\usepackage[square,numbers]{natbib}
\usepackage{blindtext}
\usepackage{enumitem}
\usepackage{xcolor}

\newcolumntype{M}[1]{>{\centering\arraybackslash}m{#1}}
\newcolumntype{N}{@{}m{0pt}@{}}

\newcommand{\BEAS}{\begin{eqnarray*}}
\newcommand{\EEAS}{\end{eqnarray*}}
\newcommand{\BEA}{\begin{eqnarray}}
\newcommand{\EEA}{\end{eqnarray}}
\newcommand{\BEQ}{\begin{equation}}
\newcommand{\EEQ}{\end{equation}}
\newcommand{\BIT}{\begin{itemize}}
\newcommand{\EIT}{\end{itemize}}
\newcommand{\BNUM}{\begin{enumerate}}
\newcommand{\ENUM}{\end{enumerate}}
\setlist[description]{leftmargin=\parindent,labelindent=\parindent}

\newcommand{\cmmnt}[1]{}

\title{A Simple Baseline Algorithm for Graph Classification}

\author{
  Nathan de Lara \\
  \texttt{Telecom ParisTech}\\
  \texttt{ndelara@enst.fr}
  \And
  Edouard Pineau \\
  \texttt{Telecom ParisTech - Safran}\\
  \texttt{edouard.pineau@safrangroup.com}
}

\begin{document}
\maketitle

\begin{abstract}
    Graph classification has recently received a lot of attention from various fields of machine learning e.g. kernel methods, sequential modeling or graph embedding. All these approaches offer promising results with different respective strengths and weaknesses. However, most of them rely on complex mathematics and require heavy computational power to achieve their best performance. We propose a simple and fast algorithm based on the spectral decomposition of graph Laplacian to perform graph classification and get a first reference score for a dataset. We show that this method obtains competitive results compared to state-of-the-art algorithms. 
\end{abstract}

\section{Introduction}
\label{sec:introduction}

Graph classification methods can schematically be divided into three categories: graph kernels, sequential methods and embedding methods. In this section, we briefly present these different approaches, focusing on methods that only use the structure of the graph and no exogenous information, such as node features, to perform classification as we only want to compare the capacity of the algorithms to capture structural information.

\paragraph{Kernel methods}
Kernel methods \citep{nikolentzos2017kernel, nikolentzos2017matching, nikolentzos2018degeneracy, neumann2016propagation} perform pairwise comparisons between the graphs of the dataset and apply a classifier, usually a support vector machine (SVM), on the similarity matrix. In order to maintain the number of comparisons tractable when the number of graphs is large, they often use Nyström algorithm \citep{williams2001using} to compute a low rank approximation of the similarity matrix. The key is to construct an efficient kernel that can be applied to graphs of varying sizes and captures useful features for the downstream classification.

\paragraph{Sequential methods}
Some methods tackle the varying sizes of graphs by processing them as a sequence of nodes. Earliest models used random walk based representations \citep{callut2008classification, xu2012protein}. More recently, \cite{jin2018learning} or \cite{you2018graphrnn} transform a graph into a sequence of fixed size vectors, corresponding to its nodes, which is fed to a recurrent neural network. The two main challenges in this approach are the design of the embedding function for the nodes and the order in which the embeddings are given to the recurrent neural network.

\paragraph{Embedding methods}
Embedding methods \citep{gomez2017dynamics, barnett2016feature, dutta2017high, DBLP:journals/corr/NarayananCVCLJ17}, derive a fixed number of features for each graph which is used as a vector representation for classification. Even though deriving a good set of features is often a difficult task, this approach has the benefit of being compatible with any standard classifier in a \textit{plug and play} fashion (SVM, random forest, multilayer perceptron...). Our model belongs to this class of methods as we rely on spectral features of the graph.

\section{Model}
\label{sec:model}


Let $G=(V, E)$ be an undirected and unweighted graph and $A \in \{0,1\}^{|V| \times |V|}$ its boolean adjacency matrix with respect to an arbitrary indexing of the nodes. $G$ is assumed to be connected, otherwise, we extract its largest connected component. Let $D = \text{diag}(A\mathbf{1})$ be the matrix of node degrees, the normalized Laplacian of $G$ is defined as

\begin{equation}
\label{eq:laplacian}
    \mathcal{L} = I - D^{-1/2}AD^{-1/2}.
\end{equation}

We use the $k$ smallest positive eigenvalues of $\mathcal{L}$ in ascending order as input of the classifier:

\begin{equation*}
    X = (\sigma_1, \dots, \sigma_k).
\end{equation*}

If the graph has less than $k$ nodes, we use right zero padding to get a vector of appropriate dimensions: $X = (\sigma_1, \dots, \sigma_{|V|-1}, 0, \dots, 0)$. We denote this embedding as spectral features (SF).

The normalized Laplacian matrix of a graph is a well-known object in spectral learning \citep{belkin2002laplacian, kamvar2003spectral}. However, for node clustering or classification most of the attention is usually directed to its eigenvectors and not its spectrum. A major benefit of the ordered spectrum representation for graph classification is that it does not depend on the indexing of the nodes. 


\paragraph{Some Laplacian eigenvalues properties}
The eigenvalues of the normalized Laplacian matrix lie between $0$ and $2$. Such a property is very convenient for the downstream use of a standard classifier without heavy rescaling or preprocessing. The multiplicity of the eigenvalue $0$ corresponds to the number of connected components in the graph, hence the omission of $\sigma_0$ in our representation as we only consider the largest connected component. Other values are also known to denote the presence of specific structures in the graph \citep{chung1997spectral}. For example, an eigenvalue equal to $2$ denotes a bipartite structure.

\paragraph{Physical interpretations} In \citep{bonald2018weighted}, each eigenvalue of the Laplacian corresponds to the energy level of a \textit{stable} configuration of the nodes in the embedding space. The lower the energy, the stabler the configuration. In \citep{shuman2016vertex}, these eigenvalues correspond to frequencies associated to a Fourier decomposition of any signal living on the vertices of the graph. Thus, the truncation of the Fourier decomposition acts as low-pass filter on the signal. Characterizing a graph by the smallest eigenvalues of its normalized Laplacian is thus comparable to characterizing a melody by its lowest fundamental frequencies.

Finally, there have been some attempts to connect spectral decomposition to graph isomorphism \citep{van2003graphs, kolla2017spectral}, however, to the best of our knowledge, this is still an open problem.

The choice of the classifier is left to the discretion of the user. In our experiments, we chose a random forest classifier (RFC) which offers a good computational speed versus accuracy trade-off. Results with several other common classifiers are displayed in appendix \ref{app:all_results}. 

\begin{figure}
    \centering
    \includegraphics[width=1\textwidth]{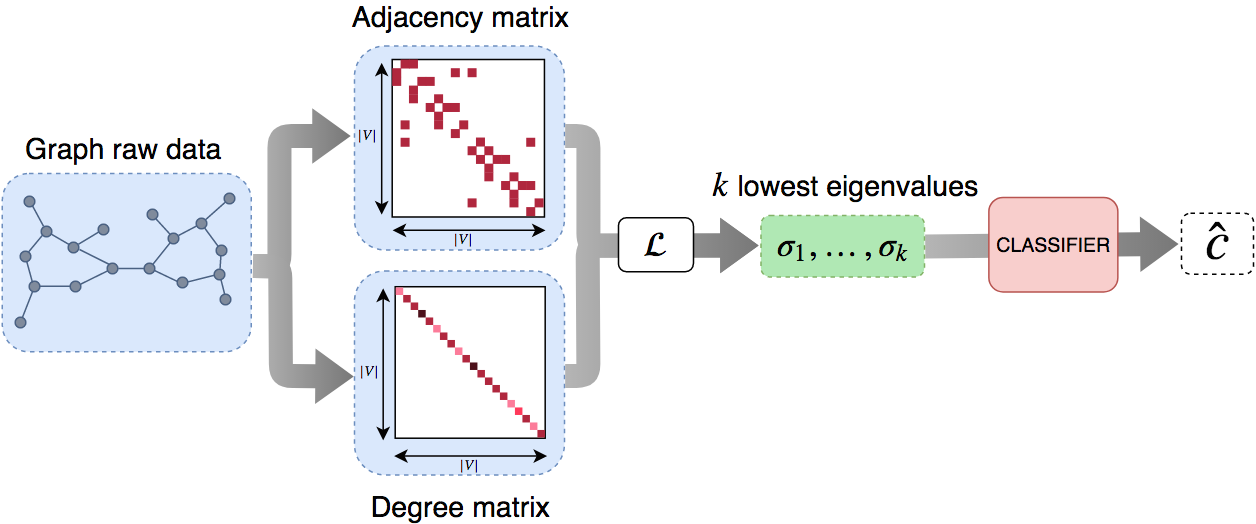}
    \caption{Schematic view of our model. $\mathcal{L}$ denotes the Laplacian as in equation \ref{eq:laplacian} and $\widehat{c}$ the predicted class.}
    \label{fig:model}
\end{figure}

An illustration of the model is proposed in figure \ref{fig:model}.

\section{Experiments}
\label{sec:experiments}

\paragraph{Datasets}
We evaluated our model against some standard datasets from biology: Mutag (MT), Predictive Toxicology Challenge (PTC), Enzymes (EZ), Proteins Full (PF), Dobson and Doig (DD) and National Cancer Institute (NCI1) \citep{KKMMN2016}. All graphs represent chemical compounds. Nodes are molecular substructures (typically atoms) and edges represent connections between these substructures (chemical bound or spatial proximity). In MT, the compounds are either mutagenic and not mutagenic while in PTC, they are either carcinogens or not. EZ contains tertiary structures of proteins from the 6 Enzyme Commission top level classes. In DD, graphs represent secondary structures of proteins being either enzyme or not enzyme. PF is a subset of DD where the largest graphs have been removed. In NCI1, compounds have either an anti-cancer activity or not. Statistics about the graphs are presented in table \ref{tab:datasets}.

\begin{table}
    \centering
    \renewcommand{\arraystretch}{1.2}
    \begin{tabular}[t]{l|cccccc}
    
                          & MT    & PTC   & EZ    & PF    & DD    & NCI1  \\
        \hline
        $\#$ graphs       & 188   & 344   & 600   & 1113  & 1178  & 4110  \\
        $\#$ classes      & 2     & 2     & 6     & 2     & 2     & 2     \\
        bias ($\%$)       & 66.5  & 55.8  & 16.7  & 59.6  & 58.7  & 50.0  \\
        avg. |V|          & 18    & 14    & 33    & 39    & 284   & 30  \\
        avg. |E|          & 39    & 15    & 124   & 146   & 1431  & 65  \\
    
    \end{tabular}
    \vspace{0.5cm}
    \caption{Basic characteristics of the datasets. Bias indicates the proportion of the dominant class.}
    \label{tab:datasets}
\end{table}

\paragraph{Experimental setup}
\label{para:experimental_setup}
Each dataset is divided into 10 folds such that the class proportions are preserved in each fold for all datasets. These folds are then used for cross-validation i.e, one fold serves as the testing set while the other ones compose the training set. Results are averaged over all testing sets. We built the folds using scikit-learn \cite{pedregosa2011scikit} \textit{StratifiedKFold} function with the random seed fixed to $1$ in order to get reproducible results.

The embedding dimension is set to the average number of nodes for each dataset (see appendix \ref{app:k_results} for additional experiments) and a unique set of hyper-parameters for the classifier is used for all datasets. We used the random forest classifier from scikit-learn with \textit{class\_weights}: balanced. The other non-default hyper parameters were selected by randomized cross validation over the different datasets (see table \ref{tab:grid_para} for more details). We also conducted experiments to ensure the robustness of our model with respect to some of its hyper-parameters, see Appendix \ref{app:robustness} for more details. All experiments were run on a laptop equipped with an intel core i7 vPro processor and 16GB of RAM.

\paragraph{Results}
We compare our results (RFC) to those obtained by Earth Mover's Distance \citep{nikolentzos2017matching} (EMD), Pyramid Match \citep{nikolentzos2017matching} (PM), Feature-Based \citep{barnett2016feature} (FB), Dynamic-Based Features \citep{gomez2017dynamics} (DyF) and Stochastic Graphlet Embedding \citep{dutta2017high} (SGE). All values are directly taken from the aforementioned papers as they used a setup similar to ours. For algorithms presenting results with and without node features, we reported the results without node features. For algorithms presenting results with several sets of hyper-parameters, we reported the results for the set of parameters that gave the best performance on the largest number of datasets. Results are reported in table \ref{tab:results}.

\begin{table}
  \begin{center}
  \renewcommand{\arraystretch}{1.2}
    \begin{tabular}[t]{l|M{1.2cm} M{1.2cm} M{1.2cm} M{1.2cm} M{1.2cm} M{1.2cm}}
                    & MT   &     PTC       &      EZ       & PF      & DD      & NCI1 \\
      \hline
      EMD           & 86.1 &     57.7      & 36.8          & -       &   -     & 72.7 \\
      PM            & 85.6 &     59.4      & 28.2          & -       &  75.6   & 69.7 \\
      FB            & 84.7 &     55.6      & 29.0          & 70.0    &   -     & 62.9 \\
      DyF           & 86.3 &     56.2      & 26.6          & 73.1    &   -     & 66.6 \\
      SGE           & 87.2 & 60.0 & 40.7          & -    &  \textbf{76.6} & -\\
      \hline
      SF + RFC      & \textbf{88.4} & \textbf{62.8} & \textbf{43.7} & \textbf{73.6} &  75.4   & \textbf{75.2} \\
      
      \end{tabular}
      \vspace{0.5cm}
      \renewcommand{\arraystretch}{4}
  \end{center}
  \caption{Experimental accuracy ($\%$) of different models plus ours over standard molecular datasets.}
  \label{tab:results}
\end{table}

We see that our model achieves good performance compared to the state-of-the art. It gives the best result on five out of the six datasets (MT, PTC, EZ, PF, NCI1). Besides, it did not require any per-dataset hyper parameters intensive tuning as we used the same random forest for all datasets.

\paragraph{Computation analysis}

The results were obtained extremely quickly (some kernel methods cannot run within one day on DD for example \citep{dutta2017high}). Embedding all graphs took approximately $3$ minutes
(most of it dedicated to DD which has the largest graphs and largest embedding dimension), while training and testing the random forest on all $60$ folds took less than a minute. Hence, the total time to run all described experiments was less than $5$ minutes.

\paragraph{Conclusion} We experimentally showed the interest of normalized Laplacian eigenvalues for graph classification. This feature is easy to extract and can be combined to any other graph representation in order to improve the model performances. We hope it will inspire new approaches to graph classification. Experimenting with permutation-invariant classifiers \citep{lucas2018mixed, qi2017pointnet} could be a natural continuation of this work in order to properly include information from eigenvectors of $\mathcal{L}$ which are node-indexing dependent.

\paragraph{Acknowledgments}
We would like to thank Thomas Bonald, Sebastien Razakarivony and all the anonymous reviewers for their comments and help. This work is supported by the company Safran through the CIFRE convention 2017/1317.

\clearpage
\appendix

\section{Results for different classifiers}
\label{app:all_results}

Besides RFC, we experimented with different standard classifiers combined to our spectral embedding. Namely: $k$-nearest neighbors classifier ($k$-NNC), 2-layers perceptron with Relu non-linearity (MLP), support vector machine with \textit{one versus one} classification (SVM) and ridge regression classifier (RRC). Results are reported in table \ref{tab:all_results}.

\begin{table}[h]
  \begin{center}
  \renewcommand{\arraystretch}{1.2}
    \begin{tabular}[t]{l|M{1.2cm} M{1.2cm} M{1.2cm} M{1.2cm} M{1.2cm} M{1.2cm}}
                    & MT   &     PTC       &      EZ       & PF      & DD      & NCI1 \\
    \hline
      SF + RFC    & 88.4 & 62.8 & 43.7 & 73.6 &  75.4   & 75.2 \\
      SF + 1-NNC  & 86.8 & 59.3 & 37.3 & 65.6 &  69.6   & 68.3 \\
      SF + 15-NNC & 85.7 & 61.9 & 33.7 & 70.4 &  75.0   & 69.6 \\
      SF + MLP    & 86.3 & 60.5 & 31.8 & 71.6 &  75.6   & 62.3 \\
      SF + SVM    & 85.3 & 60.8 & 31.3 & 73.0 &  75.0   & 63.9 \\
      SF + RRC    & 84.2 & 59.6 & 26.7 & 71.5 &  75.0   & 62.2 \\
      
      \end{tabular}
      \renewcommand{\arraystretch}{4}
  \end{center}
  \caption{Accuracy ($\%$) of different classifiers combined to the spectral features embedding.}
  \label{tab:all_results}
\end{table}


As we can see, RFC provides the best results for all datasets except DD where MLP has an accuracy of 75.6 against 75.4. Our intuition to explain these good results is that the decision tree classifier, which is at the core of RFC, is an algorithm based on level thresholding. As explained in section \ref{sec:model}, our embedding represents a sequence of energy levels, being above or below a certain level is thus likely to be meaningful for classification.

\section{Results for different embedding dimensions}
\label{app:k_results}

We experimented with different embedding dimensions for RFC: $k \in \{1, 5, 10, 25, 50 \}$. The hyperparameters are the same as in section \ref{sec:experiments}. Results are reported in table \ref{tab:k_results}.

\begin{table}[h]
  \begin{center}
  \renewcommand{\arraystretch}{1.2}
    \begin{tabular}[t]{l|M{1.2cm} M{1.2cm} M{1.2cm} M{1.2cm} M{1.2cm} M{1.2cm}}
         $k$       & MT   &     PTC       &      EZ       & PF      & DD      & NCI1 \\
    \hline
      1  & 76.2 & 56.1 & 23.8 & 64.0 &  57.2   & 58.2 \\
      5  & 86.8 & 62.5 & 39.0 & 69.6 &  73.9   & 72.5 \\
      10 & 86.8 & 61.4 & 42.8 & 71.7 &  75.5   & 75.5 \\
      25 & 88.4 & 62.8 & 42.7 & 72.8 &  75.7   & 75.2 \\
      50 & 88.4 & 62.8 & 43.7 & 73.6 &  75.1   & 75.2 \\
      
      \end{tabular}
      \renewcommand{\arraystretch}{4}
  \end{center}
  \caption{Accuracy ($\%$) of RF combined to the spectral features embedding of different dimensions.}
  \label{tab:k_results}
\end{table}

We see that even the first energy level is sufficient to obtain a non-trivial classification. $k=5$ provides results competitive with the state of the art while $k=50$ provides results relatively similar to $k= avg(|V|)$. We did not experiment with larger values of $k$ as it would mostly result into additional zero padding for most graphs. Note that embedding all graphs for $k=50$ took less than a minute in our experimental setting.

\section{Hyper parameters search and robustness analysis}
\label{app:robustness}

\begin{figure}
    \begin{center}
        \includegraphics[width=0.49\textwidth]{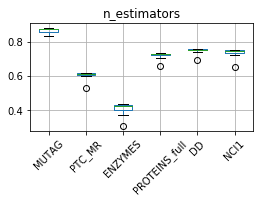}
        \includegraphics[width=0.49\textwidth]{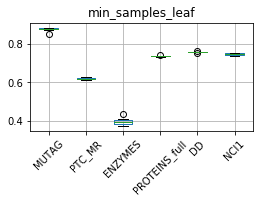} \\
        \includegraphics[width=0.49\textwidth]{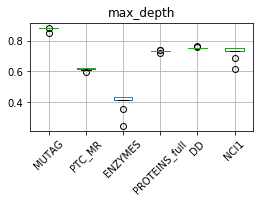}
        \includegraphics[width=0.49\textwidth]{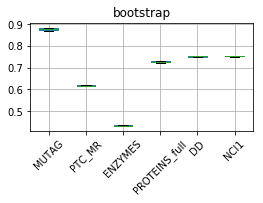} \\
    \end{center}

    \caption{Box plot representing $25^{th}$, $50^{th}$ and $75^{th}$ percentiles (box), confidence (moustaches) and outliers (isolated points) of the empirical distribution of the classification accuracy with respect to four hyperparameters of the RFC.}
    \label{fig:hp_sensitivity}
\end{figure}

In order to confirm the intrinsic quality of our spectral graph representation, we performed robustness analysis of our model with respect to the classifier. To do so, we measured the marginal variation of accuracy with respect to some hyperparameters, the others being fixed.
To ensure that we only capture parameters sensibility, we fixed the seed of the random forest to $1$ for all experiments. See table \ref{tab:grid_para} for the parameters grid and figure \ref{fig:hp_sensitivity} for the results.

We see that our method is very robust against RFC hyperparameters variability. Outliers in boxplots are all due to highly improper parameters ($n\_estimators=1$, $max\_depth=1$, \dots).

\begin{table}
    \centering
    \renewcommand{\arraystretch}{1.2}
    \begin{tabular}[t]{l|l}
    
        RFC hyperparameters     & Hyperparameters grid \\
        \hline
        $n\_estimators$         &1, 10, 50, 100, 250, \textbf{500}, 750, 1000      \\
        $min\_samples\_leaf$    & \textbf{1}, 2, 3, 4, 5, 6 \\
        $max\_depth$            & 1, 5, 10, 50, \textbf{100}, 250, 500, 750, 1000 \\
        $boostrap$              & \textbf{True}, False \\

    \end{tabular}
    
    \caption{Parameters grid for RFC. Bold values correspond to parameters used in the experimental setup and reference values for robustness analysis.}
    \label{tab:grid_para}
\end{table}

\begin{thebibliography}{10}

\bibitem{barnett2016feature}
Ian Barnett, Nishant Malik, Marieke~L Kuijjer, Peter~J Mucha, and Jukka-Pekka
  Onnela.
\newblock Feature-based classification of networks.
\newblock {\em arXiv preprint arXiv:1610.05868}, 2016.

\bibitem{belkin2002laplacian}
Mikhail Belkin and Partha Niyogi.
\newblock Laplacian eigenmaps and spectral techniques for embedding and
  clustering.
\newblock In {\em Advances in neural information processing systems}, pages
  585--591, 2002.

\bibitem{bonald2018weighted}
Thomas Bonald, Alexandre Hollocou, and Marc Lelarge.
\newblock Weighted spectral embedding of graphs.
\newblock {\em arXiv preprint arXiv:1809.11115}, 2018.

\bibitem{callut2008classification}
J{\'e}r{\^o}me Callut, Kevin Fran{\c{c}}oisse, Marco Saerens, and Pierre
  Dupont.
\newblock Classification in graphs using discriminative random walks, 2008.

\bibitem{chung1997spectral}
Fan~RK Chung and Fan~Chung Graham.
\newblock {\em Spectral graph theory}.
\newblock Number~92. American Mathematical Soc., 1997.

\bibitem{dutta2017high}
Anjan Dutta and Hichem Sahbi.
\newblock High order stochastic graphlet embedding for graph-based pattern
  recognition.
\newblock {\em arXiv preprint arXiv:1702.00156}, 2017.

\bibitem{gomez2017dynamics}
Leonardo~Gutierrez Gomez, Benjamin Chiem, and Jean-Charles Delvenne.
\newblock Dynamics based features for graph classification.
\newblock {\em arXiv preprint arXiv:1705.10817}, 2017.

\bibitem{jin2018learning}
Yu~Jin and Joseph~F JaJa.
\newblock Learning graph-level representations with gated recurrent neural
  networks.
\newblock {\em arXiv preprint arXiv:1805.07683}, 2018.

\bibitem{kamvar2003spectral}
Kamvar Kamvar, Sepandar Sepandar, Klein Klein, Dan Dan, Manning Manning, and
  Christopher Christopher.
\newblock Spectral learning.
\newblock In {\em International Joint Conference of Artificial Intelligence}.
  Stanford InfoLab, 2003.

\bibitem{KKMMN2016}
Kristian Kersting, Nils~M. Kriege, Christopher Morris, Petra Mutzel, and Marion
  Neumann.
\newblock Benchmark data sets for graph kernels, 2016.
\newblock \url{http://graphkernels.cs.tu-dortmund.de}.

\bibitem{kolla2017spectral}
Alexandra Kolla, Yannis Koutis, Vivek Madan, and Ali~Kemal Sinop.
\newblock Spectral graph isomorphism.
\newblock 2017.

\bibitem{lucas2018mixed}
Thomas Lucas, Corentin Tallec, Jakob Verbeek, and Yann Ollivier.
\newblock Mixed batches and symmetric discriminators for gan training.
\newblock {\em arXiv preprint arXiv:1806.07185}, 2018.

\bibitem{DBLP:journals/corr/NarayananCVCLJ17}
Annamalai Narayanan, Mahinthan Chandramohan, Rajasekar Venkatesan, Lihui Chen,
  Yang Liu, and Shantanu Jaiswal.
\newblock graph2vec: Learning distributed representations of graphs.
\newblock {\em CoRR}, abs/1707.05005, 2017.

\bibitem{neumann2016propagation}
Marion Neumann, Roman Garnett, Christian Bauckhage, and Kristian Kersting.
\newblock Propagation kernels: efficient graph kernels from propagated
  information.
\newblock {\em Machine Learning}, 102(2):209--245, 2016.

\bibitem{nikolentzos2018degeneracy}
Giannis Nikolentzos, Polykarpos Meladianos, Stratis Limnios, and Michalis
  Vazirgiannis.
\newblock A degeneracy framework for graph similarity.
\newblock In {\em IJCAI}, pages 2595--2601, 2018.

\bibitem{nikolentzos2017kernel}
Giannis Nikolentzos, Polykarpos Meladianos, Antoine Jean-Pierre Tixier,
  Konstantinos Skianis, and Michalis Vazirgiannis.
\newblock Kernel graph convolutional neural networks.
\newblock {\em arXiv preprint arXiv:1710.10689}, 2017.

\bibitem{nikolentzos2017matching}
Giannis Nikolentzos, Polykarpos Meladianos, and Michalis Vazirgiannis.
\newblock Matching node embeddings for graph similarity.
\newblock In {\em AAAI}, pages 2429--2435, 2017.

\bibitem{pedregosa2011scikit}
Fabian Pedregosa, Ga{\"e}l Varoquaux, Alexandre Gramfort, Vincent Michel,
  Bertrand Thirion, Olivier Grisel, Mathieu Blondel, Peter Prettenhofer, Ron
  Weiss, Vincent Dubourg, et~al.
\newblock Scikit-learn: Machine learning in python.
\newblock {\em Journal of machine learning research}, 12(Oct):2825--2830, 2011.

\bibitem{qi2017pointnet}
Charles~R Qi, Hao Su, Kaichun Mo, and Leonidas~J Guibas.
\newblock Pointnet: Deep learning on point sets for 3d classification and
  segmentation.
\newblock {\em Proc. Computer Vision and Pattern Recognition (CVPR), IEEE},
  1(2):4, 2017.

\bibitem{shuman2016vertex}
David~I Shuman, Benjamin Ricaud, and Pierre Vandergheynst.
\newblock Vertex-frequency analysis on graphs.
\newblock {\em Applied and Computational Harmonic Analysis}, 40(2):260--291,
  2016.

\bibitem{van2003graphs}
Edwin~R Van~Dam and Willem~H Haemers.
\newblock Which graphs are determined by their spectrum?
\newblock {\em Linear Algebra and its applications}, 373:241--272, 2003.

\bibitem{williams2001using}
Christopher~KI Williams and Matthias Seeger.
\newblock Using the nystr{\"o}m method to speed up kernel machines.
\newblock In {\em Advances in neural information processing systems}, pages
  682--688, 2001.

\bibitem{xu2012protein}
Xiaohua Xu, Lin Lu, Ping He, Zhoujin Pan, and Cheng Jing.
\newblock Protein classification using random walk on graph.
\newblock In {\em International Conference on Intelligent Computing}, pages
  180--184. Springer, 2012.

\bibitem{you2018graphrnn}
Jiaxuan You, Rex Ying, Xiang Ren, William~L Hamilton, and Jure Leskovec.
\newblock Graphrnn: A deep generative model for graphs.
\newblock {\em arXiv preprint arXiv:1802.08773}, 2018.

\end{thebibliography}
\end{document}